\def\year{2018}\relax
\documentclass[letterpaper]{article} 
\usepackage{aaai18}  
\usepackage{times}  
\usepackage{helvet}  
\usepackage{courier}  
\usepackage{url}  
\usepackage{graphicx}  
\usepackage{mathtools}
\usepackage{amsthm}
\usepackage[normalem]{ulem}
\usepackage{csquotes}
\usepackage{amssymb}
\usepackage{amsmath}

\begin{document}
%
\title{ExprGAN: Facial Expression Editing with Controllable Expression Intensity}
\author{Hui Ding,$^1$ Kumar Sricharan$^2$, Rama Chellappa$^3$\\
$^{1,3}$University of Maryland, College Park\\
$^2$PARC, Palo Alto\\
}
\maketitle
\begin{abstract}
Facial expression editing is a challenging task as it needs a high-level semantic understanding of the input face image.
In conventional methods, either paired training data is required or the synthetic face's resolution is low. Moreover, only the categories of facial expression can be changed.
To address these limitations,
we propose an Expression Generative Adversarial Network (ExprGAN) for photo-realistic facial expression editing with controllable expression intensity.
An expression controller module is specially designed to learn an expressive and compact expression code in addition to the encoder-decoder network. 
This novel architecture enables the expression intensity to be continuously adjusted from low to high.
We further show that our ExprGAN can be applied for other tasks, such as expression transfer, image retrieval, and data augmentation for training improved face expression recognition models. To tackle the small size of the training database, 
an effective incremental learning scheme is proposed.
Quantitative and qualitative evaluations on the widely used Oulu-CASIA dataset demonstrate the effectiveness of ExprGAN.

\end{abstract}

\section{Introduction}
\noindent Facial expression editing is the task that transforms the expression of a given face image to a target one without affecting the identity properties. It has applications in facial animation, human-computer interactions, entertainment, etc. The area has been attracting considerable attention both from academic and industrial research communities.

Existing methods that address expression editing can be divided into two categories. One category tries to manipulate images by reusing parts of existing ones~\cite{yang2011expression,mohammed2009visio,yeh2016semantic} while the other
resorts to synthesis techniques to generate a face image with the target 
expression~\cite{susskind2008generating,reed2014learning,cheung2014discovering}.
In the first category, traditional methods~\cite{yang2011expression} often make use of the expression flow map to transfer an expression by image warping. Recently, \citeauthor{yeh2016semantic}~\shortcite{yeh2016semantic} applies the idea to a variational autoencoder to learn the flow field. 
Although the generated face image has high resolution, 
paired data where one subject has different expressions are needed to train the model.
In the second category, deep learning-based methods are mainly used. The early work by~\citeauthor{susskind2008generating}~\shortcite{susskind2008generating} uses a deep belief network to generate emotional faces, which can be controlled by the Facial Action Coding System (FACS) labels. 
In~\cite{reed2014learning}, a three-way gated Boltzmann machine is employed to model the relationships between the expression and identity. 
However, the synthesized image of these methods has low resolution (48 x 48), lacking fine details and tending to be blurry. 

Moreover, existing works can only transform the expression to different classes, like \textit{Angry} or \textit{Happy}.
However, in reality, the intensity of facial expression is often displayed over a range. 
For example, humans can express the \textit{Happy} expression either with a huge grin or by a gentle smile. Thus it is appealing if both the type of the expression and its intensity can be controlled simultaneously. Motivated by this, 
in this paper, we present a new expression editing model, Expression Generative Adversarial Network (ExprGAN) which has the unique property that 
\textbf{\textit{multiple diverse styles of the target expression can be synthesized where the
intensity of the generated expression is able to be controlled continuously from weak to strong, without the need for training data with intensity values}}.

To achieve this goal, we specially design an expression controller module. Instead of feeding in a deterministic one-hot vector label like previous works, the expression code generated by the expression controller module is used.
It is a real-valued vector conditioned on the label,
thus more complex information such as expression intensity can be described.
Moreover, to force each dimension of the expression code to capture a different factor of the intensity variations,
the conditional mutual information between the generated image and the expression code is maximized by a regularizer network.

Our work is inspired by the recent success of the image generative model, where a generative adversarial network~\cite{goodfellow2014generative} learns to produce samples similar to a given data distribution through a two-player game between a generator and a discriminator. Our ExprGAN also adopts the generator and discriminator framework in addition to the expression controller module and regularizer network. However, to facilitate image editing, the generator is composed of an encoder and a decoder. The input of the encoder is a face image, the output of the decoder is a reconstructed one, and the learned identity and expression representations bridge the encoder and decoder. To preserve the most prominent facial structure,
we adopt a multi-layer perceptual loss~\cite{johnson2016perceptual} in the feature space in addition to the pixel-wise $L_1$ loss. 
Moreover, to make the synthesized image look more photo-realistic, two adversarial networks are imposed on the encoder and decoder, respectively. 
Because it is difficult to directly train our model on the small training set,
a three-stage incremental learning algorithm is also developed.

The main contributions of our work are as follows:
\begin{itemize}
\item We propose a novel model called ExprGAN that can change a face image to a target expression with multiple styles, where the expression intensity can also be controlled continuously.
\item We show that the synthesized face images have high perceptual quality, which can be used to improve the performance of an expression classifier.
\item Our identity and expression representations are explicitly disentangled which can be exploited for tasks such as expression transfer, image retrieval, etc. 
\item We develop an incremental training strategy to train the model on a relative small dataset without the rigid requirement of paired samples.
\end{itemize}

\begin{figure*}
\centering
\includegraphics[width=0.9\textwidth]{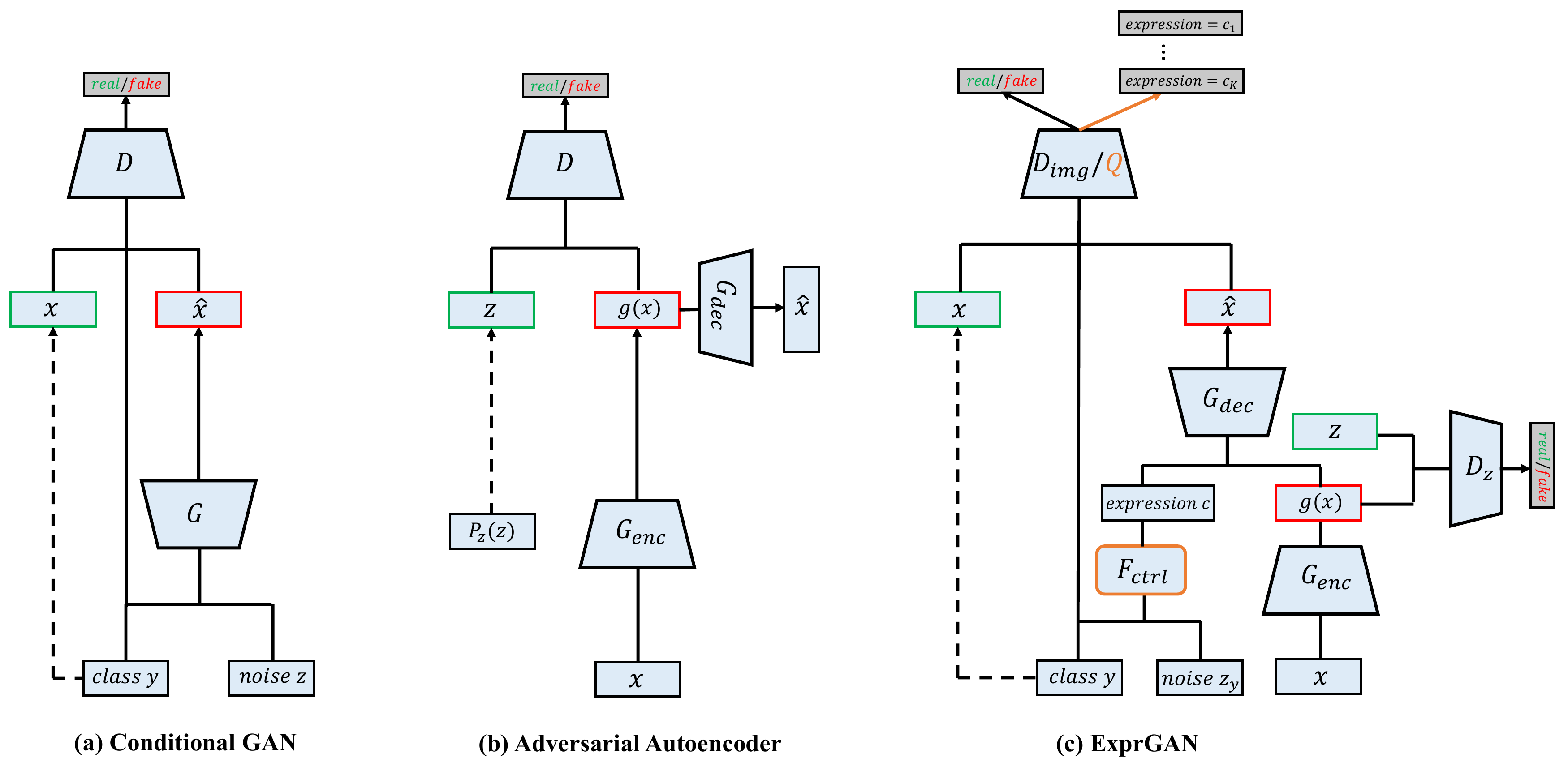}
\caption{Comparison of previous GAN architectures and our proposed ExprGAN.
}
\label{fig:framework}
\end{figure*}

\section{Related Works}
\subsubsection{Deep Generative Model}
Deep generative models have achieved impressive success in recent years.
There are two major approaches: generative adversarial network (GAN)~\cite{goodfellow2014generative} and variational autoencoder (VAE)~\cite{kingma2013auto}. GAN is composed of a generator and a discriminator, where the training is carried out with a minimax two-player game. GAN has been used for image synthesis~\cite{radford2015unsupervised}, image superresolution~\cite{ledig2016photo}, etc. One interesting extension of GAN is Conditional GAN (CGAN)~\cite{mirza2014conditional} where the generated image can be controlled by the condition variable. 
On the other hand, VAE is a probabilistic model with an encoder to map an image to a latent representation and a decoder to reconstruct the image. A reparametrization trick is proposed which enables the model to be trained by backpropogation~\cite{rumelhart1988learning}. 
One variant of VAE is Adversarial Autoencoder~\cite{makhzani2015adversarial}, where an adversarial network is adopted to regularize the latent representation to conform to a prior distribution. 
Our ExprGAN also adopts an autoencoder structure, but there are two main differences: 
First, an expression controller module is specially designed, so a face with different types of expressions across a wide range of intensities can be generated.
Second, to improve the generated image quality, a face identity preserving loss and two adversarial losses are incorporated.

\subsubsection{Facial Expression Editing}
Facial expression editing has been actively investigated in computer graphics. 
Traditional approaches include 3D model-based~\cite{blanz2003reanimating}, 2D expression mapping-based~\cite{liu2001expressive} and flow-based~\cite{yang2011expression}. 
Recently, deep learning-based methods have been proposed. 
\citeauthor{susskind2008generating}~\shortcite{susskind2008generating} studied a deep belief network to generate facial expression given high-level identity and facial action unit (AU) labels.
In~\cite{reed2014learning}, a higher-order Boltzman machine with multiplicative
interactions was proposed to model the distinct factors of variation. 
\citeauthor{cheung2014discovering}~\shortcite{cheung2014discovering} proposed a decorrelating regularizer to 
disentangle the variations between identity and expression in an unsupervised manner. 
However, the generated image is low resolution with size of 48 x 48, which is not visually satisfying.
Recently, \citeauthor{yeh2016semantic}~\shortcite{yeh2016semantic} proposed to edit the facial expression by image warping with appearance flow.
Although the model can generate high-resolution images, paired samples as well as the labeled query image are required.

The most similar work to ours is CFGAN~\cite{kaneko2017cvpr}, which uses a filter module to control the generated face attributes. However, there are two main differences: 
First, CFGAN adopts the CGAN architecture where an encoder needs to be trained separately for image editing.
While for the proposed ExprGAN, the encoder and the decoder are constructed in a unified framework. 
Second, the attribute filter of CFGAN is mainly designed for a single class, while our expression controller module works for multiple categories. 
Most recently,~\citeauthor{zhang2017age} \shortcite{zhang2017age} proposed a conditional AAE (CAAE) for face aging, which can also be applied for expression editing.
Compared with these studies, our ExprGAN has two main differences: First, in addition to transforming a given face image to a new facial expression, our model can also control the expression intensity continuously without the intensity training labels; Second, photo-realistic face images with new identities can be generated for data augmentation, which is found to be useful to train an improved expression classifier.

\section{Proposed Method}
In this section, we describe our Expression Generative Adversarial Network (ExprGAN). We first describe the Conditional Generative Adversarial Network (CGAN)~\cite{mirza2014conditional} and Adversarial Autoencoder (AAE)~\cite{makhzani2015adversarial}, which form the basis of ExprGAN. Then the formulation of ExprGAN is explained. The architectures of the three models are shown in Fig.~\ref{fig:framework}.

\subsection{Conditional Generative Adversarial Network}
CGAN is an extension of a GAN~\cite{goodfellow2014generative} for conditional image generation. It is composed of two networks: a generator network G and a discriminator network D that compete in a two-player minimax game. Network G is trained to produce a synthetic image $\hat{x} = G(z, y)$ to fool D to believe it is an actual photograph, where $z$ and $y$ are the random noise and condition variable, respectively. D tries to distinguish the real image $x$ and the generated one $\hat{x}$. Mathematically, the objective function for G and D can be written as follows:
\begin{equation}
\begin{split}
\min_G \max_D & ~\mathbb{E}_{x,y\sim P_{data}(x,y)}[\log D(x,y)] \\
&+ \mathbb{E}_{z\sim P_z(z), y\sim P_y(y)}[\log(1-D(G(z,y),y))]
\end{split}
\end{equation}

\subsection{Adversarial Autoencoder}
AAE~\cite{makhzani2015adversarial} is a probabilistic autoencoder which consists of an encoder $G_{enc}$, a decoder $G_{dec}$ and a discriminator $D$. Apart from the reconstruction loss, the hidden code vector $g(x)=G_{enc}(x)$ is also regularized by an adversarial network to impose a prior distribution $P_z(z)$. Network $D$ aims to discriminate $g(x)$ from $z\sim P_z(z)$, while $G_{enc}$ is trained to generate $g(x)$ that could fool $D$. 
Thus, the AAE objective function becomes:
\begin{equation}
\begin{split}
&\min_{G_{enc},G_{dec}}\max_D L_p(G_{dec}(G_{enc}(x)), x)\\
&+\mathbb{E}_{z\sim P_z(z)}[\log D(z)] + \mathbb{E}_{x\sim P_{data}(x)}[\log(1-D(G_{enc}(x))]
\end{split}
\end{equation}
where $L_p(,)$ is the $p_{th}$ norm: $L_p(x', x) = ||x'-x||_p^p$

\subsection{Expression Generative Adversarial Network}
Given a face image $x$ with expression label $y$, the objective of our learning problem is to edit the face to display a new type of expression at different intensities. Our approach is to train a ExprGAN conditional on the original image $x$ and the expression label $y$ with its architecture illustrated in Fig.~\ref{fig:framework} (c).

\subsubsection{Network Architecture}
ExprGAN first applies an encoder $G_{enc}$ to map the image $x$ to a latent representation $g(x)$ that preserves identity. Then, an expression controller module $F_{ctrl}$ is adopted to convert the one-hot expression label $y$ to a more expressive expression code $c$. To further constrain the elements of $c$ to capture the various aspects of the represented expression, a regularizer $Q$ is exploited to maximize the conditional mutual information between $c$ and the generated image.
Finally, the decoder $G_{dec}$ generates a reconstructed image $\hat{x}$ combining the information from $g(x)$ and $c$. To further improve the generated image quality, 
a discriminator $D_{img}$ on the decoder $G_{dec}$ is used to refine the synthesized image $\hat{x}$ to have photo-realistic textures.
Moreover, to better capture the face manifold, a discriminator $D_{z}$ on the encoder $G_{enc}$ is applied to ensure the learned identity representation is filled and exhibits no \enquote{holes}~\cite{makhzani2015adversarial}.

\subsubsection{Expression Controller Networks $F_{ctrl}$ and $Q$}
In previous conditional image generation methods~\cite{tran2017disentangled,zhang2017age}, a binary one-hot vector is usually adopted as the condition variable. This is enough for generating images corresponding to different categories. 
However, 
for our problem, 
a stronger control over the synthesized facial expression is needed:
we want to change the expression intensity in addition to generating different types of expressions.
To achieve this goal, an expression controller module $F_{ctrl}$ is designed to ensure the expression code $c$ can describe the property of the expression intensity except the category information. Furthermore, a regularizer network $Q$ is proposed to enforce the elements of $c$ to capture the multiple levels of expression intensity comprehensively.

\textbf{Expression Controller Module $F_{ctrl}$}
To enhance the description capability, $F_{ctrl}$ transforms the binary input $y$ to a continuous representation $c$ by the following operation:
\begin{equation}
c_i = F_{ctrl}(y_i, z_y) = |z_y| \cdot (2y_i - 1)~~~~i=1, 2, \dots, K
\end{equation}
where the inputs are the expression label $y \in \{0,1\}^K$ and uniformly distributed $z_y\sim U(-1,1)^d$, while the output is the expression code $c=[c_1^T,\dots,c_K^T]^T \in R^{Kd}$, 
$K$ is the number of classes. 
If the $i_{th}$ class expression is present, \textit{i.e.}, $y_i=1$, $c_i\in R^d$ is set to be a positive vector within 0 and 1, while $c_j,j\neq i$ has negative values from -1 to 0. Thus, in testing, we can manipulate the elements of $c$ to generate the desired expression type. This flexibility greatly increases the controllability of $c$ over synthesizing diverse styles and intensities of facial expressions.


\textbf{Regularizer on Expression Code $Q$}
It is desirable if each dimension of $c$ could learn a different factor of the expression intensity variations. 
Then faces with a specific intensity level can be generated by manipulating the corresponding expression code. 
To enforce this constraint,
we impose a regularization on $c$ by maximizing the conditional mutual information $I(c; \hat{x}|y)$ between the generated image $\hat{x}$ and the expression code $c$. This ensures that the expression type and intensity encoded in $c$ is reflected in the image generated by the decoder.
The direct computation of $I$ is hard since it requires the posterior $P(c|\hat{x},y)$, which is generally intractable. 
Thus, a lower bound is derived with variational inference which extends~\cite{chen2016infogan} to the conditional setting:
\begin{equation}
\begin{split}
&I(c; \hat{x}|y)\\
&= H(c|y) - H(c|\hat{x}, y)\\
&=\mathbb{E}_{\hat{x}\sim G_{dec}(g(x), c)}[\mathbb{E}_{c'\sim P(c'|\hat{x}, y)}[\log P(c'|\hat{x}, y)]] + H(c|y)\\
&=\mathbb{E}_{\hat{x}\sim G_{dec}(g(x), c)}[D_{KL}(P(\cdot|\hat{x}, y) || Q(\cdot|\hat{x}, y)) + \\
&~~~~~~\mathbb{E}_{c'\sim P(c'|\hat{x}, y)}[\log Q(c'|\hat{x}, y)]] + H(c|y)\\
&\geq \mathbb{E}_{\hat{x}\sim G_{dec}(g(x), c)}[\mathbb{E}_{c'\sim P(c'|\hat{x}, y)}[\log Q(c'|\hat{x}, y)]] + H(c|y)\\
&=\mathbb{E}_{ c\sim P(c| y),\hat{x}\sim G_{dec}(g(x), c)}[\log Q(c|\hat{x}, y)] + H(c|y)
\end{split}
\end{equation}
For simplicity, the distribution of $c$ is fixed, thus $H(c|y)$ is treated as a constant. Here the auxiliary distribution $Q$ is parameterized as a neural network, thus the final loss function is defined as follows:
\begin{equation}
\min_Q L_Q = -\mathbb{E}_{ c\sim P(c| y),\hat{x}\sim G_{dec}(g(x), c)}[\log Q(c|\hat{x}, y)]
\end{equation}

\subsubsection{Generator Network $G$}
The generator network $G=(G_{enc}, G_{dec})$ adopts the autoencoder structure where the encoder $G_{enc}$ first transforms the input image $x$ to a latent representation that preserves as much identity information as possible. After obtaining the identity code $g(x)$ and the expression code $c$, the decoder $G_{dec}$ then generates a synthetic image $\hat{x}=G_{dec}(G_{enc}(x), c)$ which should be identical as $x$. For this purpose, a pixel-wise image reconstruction loss is used:
\begin{equation}
\min_{G_{enc},G_{dec}}L_{pixel} = L_1(G_{dec}(G_{enc}(x),c), x)
\end{equation}

To further preserve the face identity between $x$ and $\hat{x}$, a pre-trained discriminative deep face model is leveraged to enforce the similarity in the feature space:
\begin{equation}
\min_{G_{enc},G_{dec}}L_{id} = \sum_l\beta_lL_1(\phi_l(G_{dec}(G_{enc}(x),c)), \phi_l(x))
\end{equation}
where $\phi_l$ are the $l_{th}$ layer feature maps of a face recognition network, and $\beta_l$ is the corresponding weight. We use the activations at the $conv1\_2$, $conv2\_2$, $conv3\_2$, $conv4\_2$ and $conv5\_2$ layer of the VGG face model~\cite{parkhi2015deep}. 

\subsubsection{Discriminator on Identity Representation $D_z$}
It is a well known fact that face images lie on a manifold~\cite{he2005face,lee2003video}.
To ensure that face images generated by interpolating between arbitrary identity representations do not deviate from the face manifold~\cite{zhang2017age}, we impose a uniform distribution on $g(x)$, forcing it to populate the latent space evenly without \enquote{holes}.
This is achieved through an adversarial training process where the training objective is:
\begin{equation}
\begin{split}
\min_{G_{enc}}\max_{D_z} L_{adv}^z=&\mathbb{E}_{z\sim P_z(z)}[\log D_z(z)]\\
& + \mathbb{E}_{x\sim P_{data}(x)}[\log(1-D_z(G_{enc}(x))]
\end{split}
\end{equation}

\subsubsection{Discriminator on Image $D_{img}$}
Similar to existing methods~\cite{huang2017beyond,tran2017disentangled}, an adversarial loss between the generated image $\hat{x}$ and the real image $x$ is further adopted to improve the photorealism:
\begin{equation}
\begin{split}
\min_{G_{enc},G_{dec}} &\max_{D_{img}}L_{adv}^{img}=\mathbb{E}_{x,y\sim P_{data}(x,y)}[\log D_{img}(x,y)] + \\
&\mathbb{E}_{x,y\sim P_{data}(x,y),z_y\sim P_{z_y}(z_y)}\\
&[\log(1-D_{img}(G_{dec}(G_{enc}(x),F_{ctrl}(z_y,y)),y))]
\end{split}
\end{equation}

\subsubsection{Overall Objective Function}
The final training loss function is a weighted sum of all the losses defined above:
\begin{equation}
\begin{split}
\min_{G_{enc},G_{dec},Q} \max_{D_{img},D_z} &L_{ExprGAN} = L_{pixel} + \lambda_1 L_{id} + \lambda_2 L_Q \\
&+ \lambda_3 L_{adv}^{img}  + \lambda_4 L_{adv}^{z} + \lambda_5 L_{tv}
\end{split}
\label{eq:all}
\end{equation}
We also impose a total variation regularization $L_{tv}$~\cite{mahendran2015understanding} on the reconstructed image to reduce spike artifacts.

\subsubsection{Incremental Training}
Empirically we find that jointly training all the subnetworks yields 
poor results as we have multiple loss functions. It is difficult for the model to learn all the functions at one time considering the small size of the dataset.
Therefore, we propose an incremental training algorithm to train the proposed ExprGAN.
Overall our incremental training strategy can be seen as a form of curriculum learning, and includes three stages: controller learning stage, image reconstruction stage and image refining stage. First, we teach the network to generate the image conditionally by training $G_{dec}$, $Q$ and $D_{img}$ where the loss function only includes $L_Q$ and $L_{adv}^{img}$. $g(x)$ is set to be random noise in this stage. After the training finishes, we then teach the network to learn the disentangled representations by reconstructing the input image with $G_{enc}$ and $G_{dec}$. To ensure that the network does not forget what is already learned, $Q$ is also trained but with a decreased weight. So the loss function has three parts: $L_{pixel}$, $L_{id}$ and $L_Q$. Finally, we train the whole network to refine the image to be more photo-realistic by adding $D_{img}$ and $D_z$ with the loss function defined in (\ref{eq:all}). 
We find in our experiments that stage-wise training is crucial to learn the desired model on the small dataset.

\section{Experiments}
We first describe the experimental setup then three main applications: expression editing with continuous control over intensity, facial expression transfer and conditional face image generation for data augmentation .

\subsection{Dataset} 
We evaluated the proposed ExprGAN on the widely used Oulu-CASIA~\cite{zhao2011facial} dataset. Oulu-CASIA has 480 image sequences taken under Dark,
Strong, Weak illumination conditions. In this experiment, only videos with Strong condition captured by a VIS camera are used. There are 80 subjects and six expressions, \textit{i.e.}, \textit{Angry}, \textit{Disgust}, \textit{Fear}, \textit{Happy}, \textit{Sad} and \textit{Surprise}. The first frame is always neutral while the last frame has the peak expression. Only the last three frames are used, and the total number of images is 1440. Training and testing sets are divided based on identity, with 1296 for training and 144 for testing. We aligned the faces using the landmarks detected from~\cite{zhang2016joint}, then cropped and resized the images to dimension of 128 x 128. Lastly, we normalized the pixel values into range of [-1, 1]. To alleviate overfitting, we augmented the training data with random flipping.

\subsection{Implementation Details}
The ExprGAN mainly builds on multiple upsampling and downsampling blocks. The upsampling block consists of the nearest-neighbor upsampling followed by a 3 x 3 stride 1 convolution. The downsampling block consists of a 5 x 5 stride 2 convolution. Specifically, $G_{enc}$ has 5 downsampling blocks where the numbers of channels are 64, 128, 256, 512, 1024 and one FC layer to get the identity representation $g(x)$. For $G_{dec}$, it has 7 upsampling blocks with 512, 256, 128, 64, 32, 16, 3 channels. $D_z$ consists of 4 FC layers with 64, 32, 16, 1 channels.  
We model $Q(c|\hat{x},y)$ as a factored Gaussian, and share many parts of $Q$ with $D_{img}$ to reduce computation cost. The shared parts have 4 downsampling blocks with 16, 32, 64, 128 channels and one FC layer to output a 1024-dim representation. Then it is branched into two heads, one for $D_{img}$ and one for $Q$. $Q$ has $K$ branches $\{Q_i\}_{i=1}^K$ where each $Q_i$ has two individual FC layers with 64, $d$ channels to predict the expression code $c_i$. 
Leaky ReLU~\cite{maas2013rectifier} and batch normalization~\cite{ioffe2015batch} are applied to $D_{img}$ and $D_z$, while ReLU~\cite{krizhevsky2012imagenet} activation is used in $G_{enc}$ and $G_{dec}$. 
The random noise $z$ is uniformly distributed from -1 to 1. We fixed the dimensions of $g(x)$ and $c$ to be 50 and 30, and found this configuration sufficient for representing the identity and expression variations.

We train the networks using the Adam optimizer~\cite{kingma2014adam}, with learning rate of 0.0002, $\beta_1=0.5$, $\beta_2=0.999$ and mini-batch size of 48. 
In the image refining stage, we empirically set $\lambda_1=1$, $\lambda_2 = 1$, $\lambda_3 = 0.01$, $\lambda_4 = 0.01$, $\lambda_5 = 0.001$. 
The model is implemented using Tensorflow~\cite{abadi2016tensorflow}.
\begin{figure}[!ht]
   \centering
     \includegraphics*[width=\linewidth]{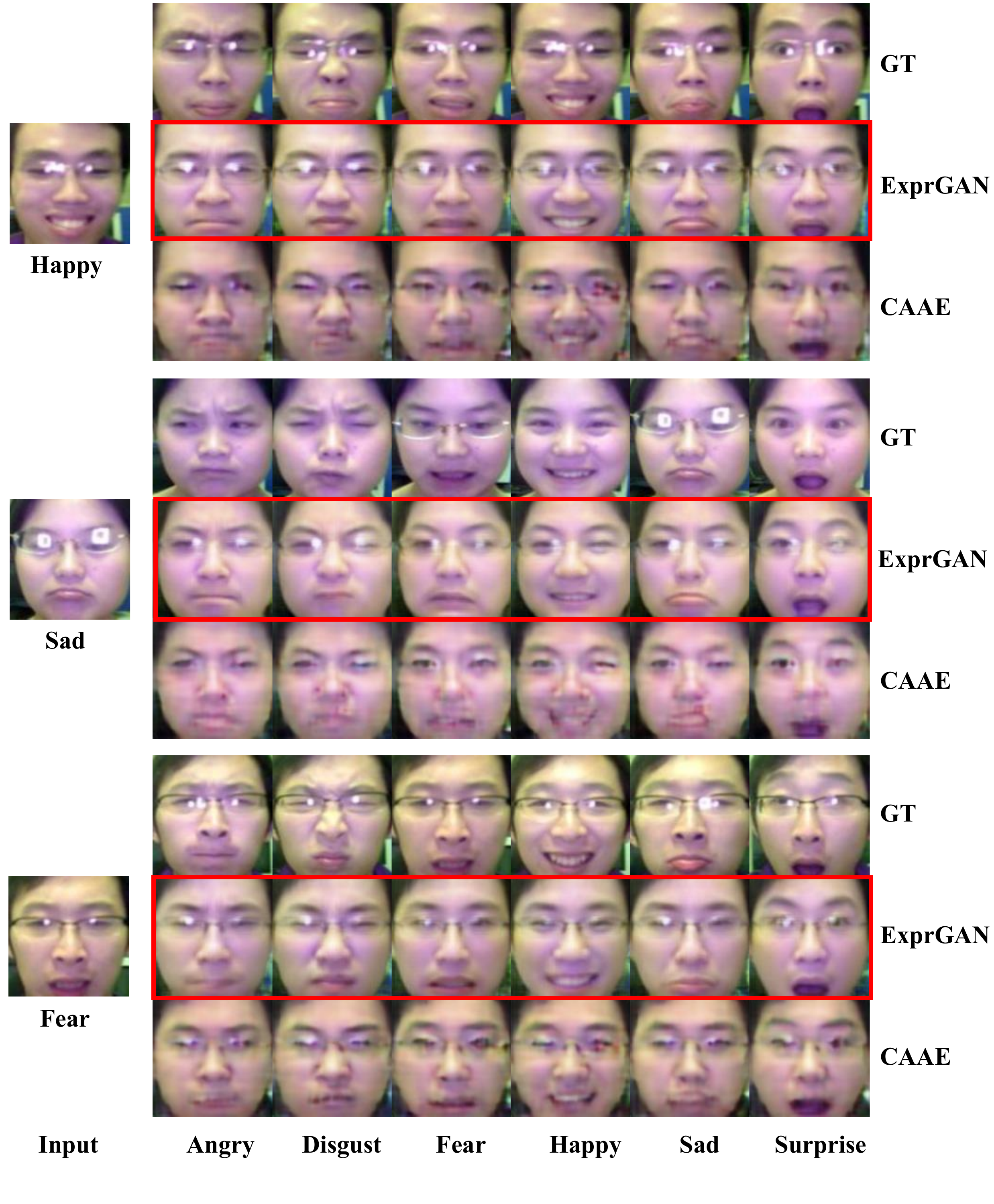}
   \caption{Visual comparison of facial expression editing results. For each input, we compare the ground truth images (top), the synthetic images of ExprGAN (middle) and CAAE (bottom). Zoom in for details.
   }
   \label{fig:edit}
\end{figure}
\subsection{Facial Expression Editing}
In this part, we demonstrate our model's ability to edit the expression of a given face image. To do this, we first input the image to $G_{enc}$ to obtain an identity representation $g(x)$. Then with the decoder $G_{dec}$, a face image of the desired expression $i$ can be generated by setting $c_i$ to be positive and $c_j,j\neq i$ to be negative. A positive (negative) value indicates the represented expression is present (absent). Here 1 and -1 are used.
Some example results are shown in Fig.~\ref{fig:edit}. The left column contains the original input images, while the middle row in the right column contains the synthesized faces corresponding to six different expressions. For comparison, the ground truth images and the results from the recent proposed CAAE~\cite{zhang2017age} are also shown in the first and third row, respectively.
We can see faces generated by our ExprGAN preserve the identities well.  
Even some subtle details like the transparent eyeglasses are also kept. Moreover, the synthesized expressions look natural. While CAAE failed to transform the input faces to new expressions with fine details, and the generated faces are blurry.

\begin{figure*}
\centering
\includegraphics[width=\textwidth]{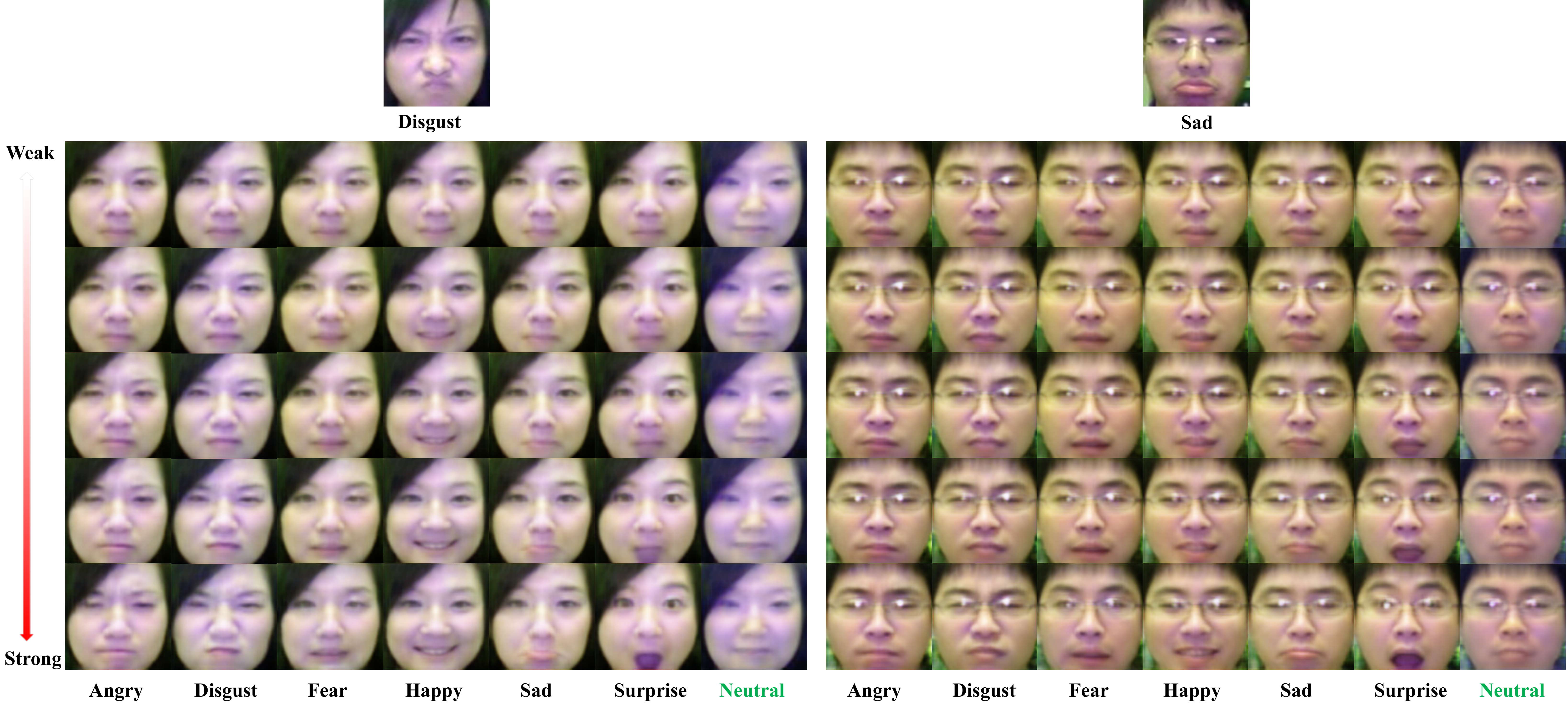} 
\caption{Face images are transformed to new expressions with different intensity levels. The top row contains the input faces with the original expressions, and the rest rows are the synthesized results.
Each column corresponds to a new expression with five intensity levels from weak to strong. The \textit{Neutral} expression which is not in the training data is also able to be generated. 
}
\label{fig:smile_disgust}
\end{figure*}

We now demonstrate that our model can transform a face image to new types of expressions with continuous intensity. This is achieved by exploiting the fact that each dimension of the expression code captures a specific level of expression intensity. In particular, to vary the intensity of the desired class $i$, we set the individual element of the expression code $c_i$ to be 1, while the other dimensions of $c_i$ and all other $c_j,j\neq i$ to be -1.
The generated results are shown in Fig.~\ref{fig:smile_disgust}. 
Take the \textit{Happy} expression in the forth column as an example. The face in the first row which corresponds to the first element of $c_i$ being 1 displays a gentle smile with mouth closed, while a big smile with white teeth is synthesized in the last row that corresponds to the fifth element of $c_i$ being 1. Moreover, when we set all $c_i$ to be -1, a \textit{Neutral} expression is able to be generated even though this expression class is not present in the training data. This validates that the expression code discovers the diverse spectrum of expression intensity in an unsupervised way, \textit{i.e.}, without the training data containing explicit labels for intensity levels.

\subsection{Facial Expression Transfer}
\begin{figure}[!ht]
   \centering
     \includegraphics*[width=\linewidth]{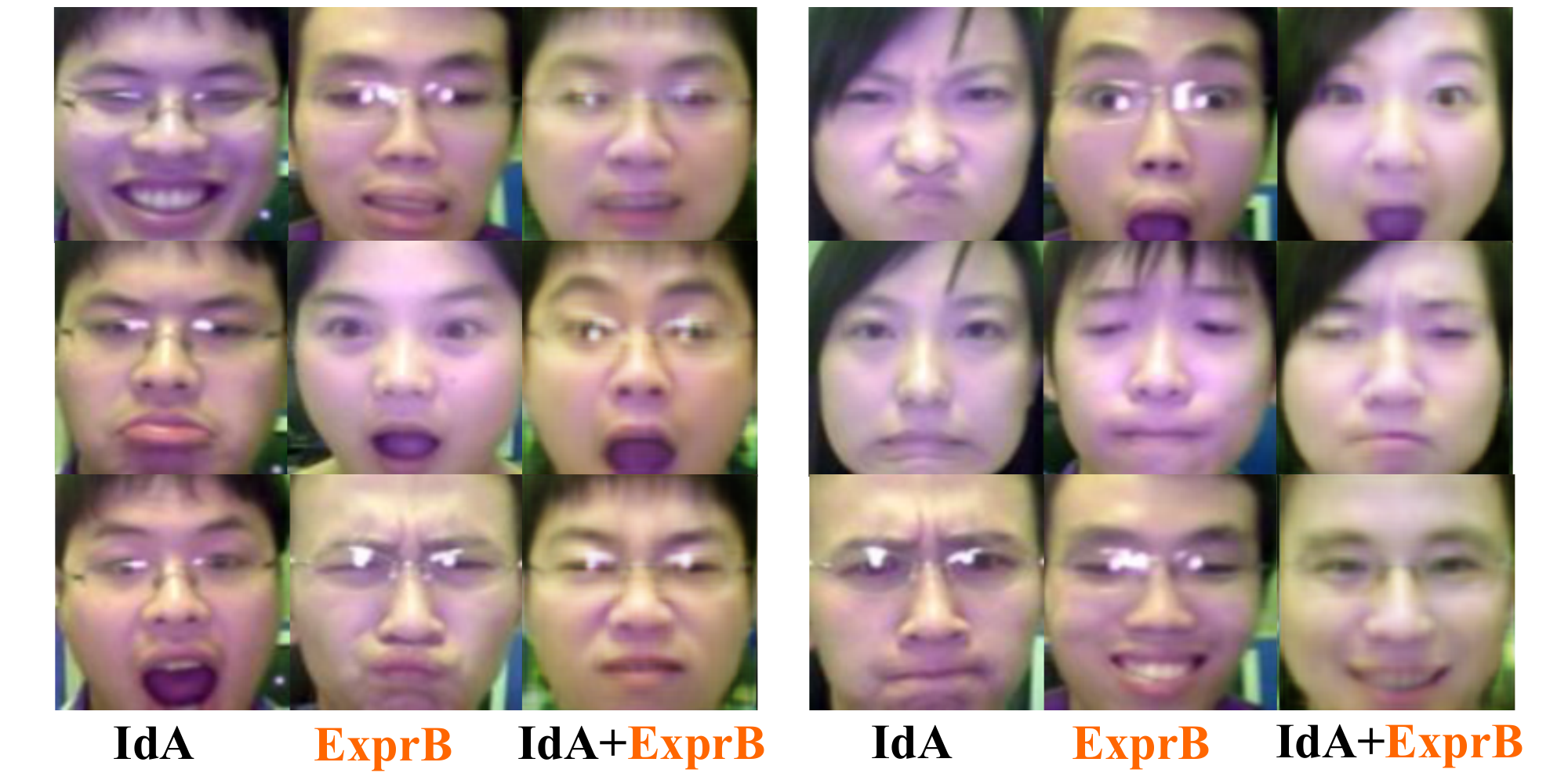}
   \caption{Facial expression transfer. Expressions from the middle column are transferred to faces in the left column. The results are shown in the right column.
   }
   \label{fig:transfer}
\end{figure}

In this part, we demonstrate our model's ability to transfer the expression of another face image $x_B$ to a given face image $x_A$. 
To do this, we first input $x_A$ to $G_{enc}$ to get the identity representation $g(x_A)$. Then we train an expression classifier to predict the expression label $y_B$ of $x_B$. With $y_B$ and $x_B$, the expression code $c_B$ can be obtained from $Q$. Finally, we can get an image with identity A and expression B from $G_{dec}(g(x_A), c_B)$. 
The generated images are shown in Fig.~\ref{fig:transfer}. 
We observe that faces having the source identities and expressions similar to the target ones can be synthesized even for some very challenging cases.
For example, when the expression \textit{Happy} is transferred to an \textit{Angry} face, the teeth region which does not exist in the source image is also able to be generated. 

\begin{figure}[!ht]
   \centering
     \includegraphics*[width=\linewidth]{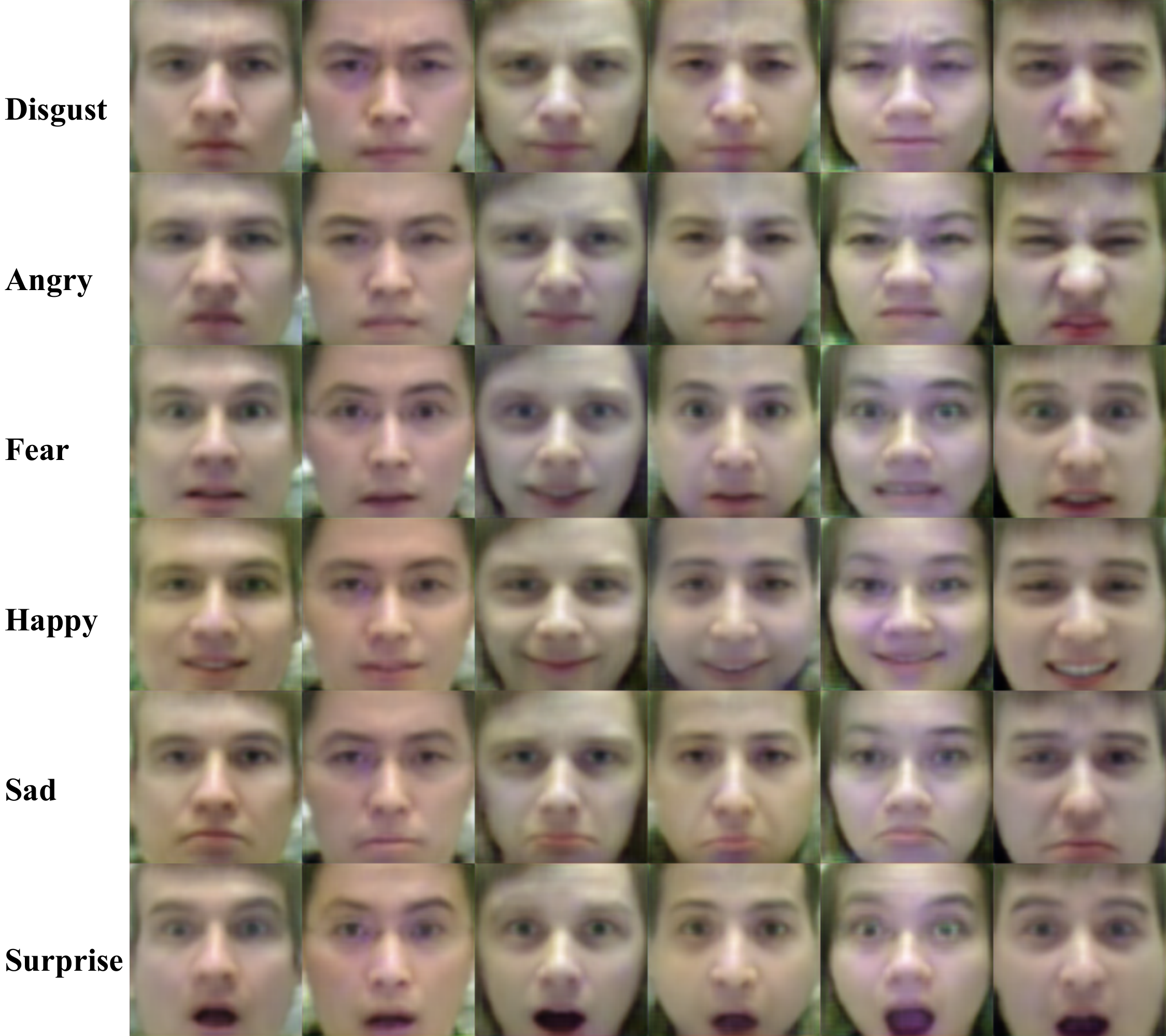}
   \caption{Random generated subjects displaying six categories of expressions. }
   \label{fig:aug}
\end{figure}
\subsection{Face Image Generation for Data Augmentation}
In this part, we first show our model's ability to generate high-quality face images controlled by the expression label, then quantitatively demonstrate the usefulness of the synthesized images.
To generate faces with new identities, we feed in random noise and expression code to $G_{dec}$.
The results are shown in Fig.~\ref{fig:aug}. Each column shows the same subject displaying different expressions. We can see the synthesized face images look realistic. Moreover, because of the design of the expression controller module, the generated expressions for the same class are also diverse. For example, for the class \textit{Happy}, there are big smile with teeth and slight smile with mouth closed.

We further demonstrate that the images synthesized by our model can be used for data augmentation to train a robust expression classifier. Specifically, for each expression category, we generate 0.5$K$, 1$K$, 5$K$, and 10$K$ images, respectively. The classifier has the same network architecture as $G_{enc}$ except one additional FC layer with six neurons is added. 
The results are shown in Table~\ref{tab:acc}. We can see by only adding 3$K$ synthetic images, the improvement is marginal, with an accuracy of 78.47\% vs. 77.78\%. However, when the number is increased to 30$K$, the recognition accuracy is improved significantly, reaching to \textbf{84.72\%} with a relative error reduction by \textbf{31.23\%}. The performance starts to saturate when more images (60$K$) are utilized.
This validates the synthetic face images have high perceptual quality.
\begin{table}
\caption{Comparison of expression recognition accuracy with different numbers of synthesized images.}
\label{tab:acc}
\centering
\begin{tabular}{c|c|c|c|c|c}
\hline
\hline
\# Syn. Images&0 & 3$K$ & 6$K$ & 30$K$ & 60$K$\\
\hline
Accuracy (\%)&77.78&78.47&81.94&\textbf{84.72}&84.72\\
\hline
\end{tabular}
\end{table}

\subsection{Feature Visualization}
\begin{figure}[!ht]
   \centering
     \includegraphics*[width=\linewidth]{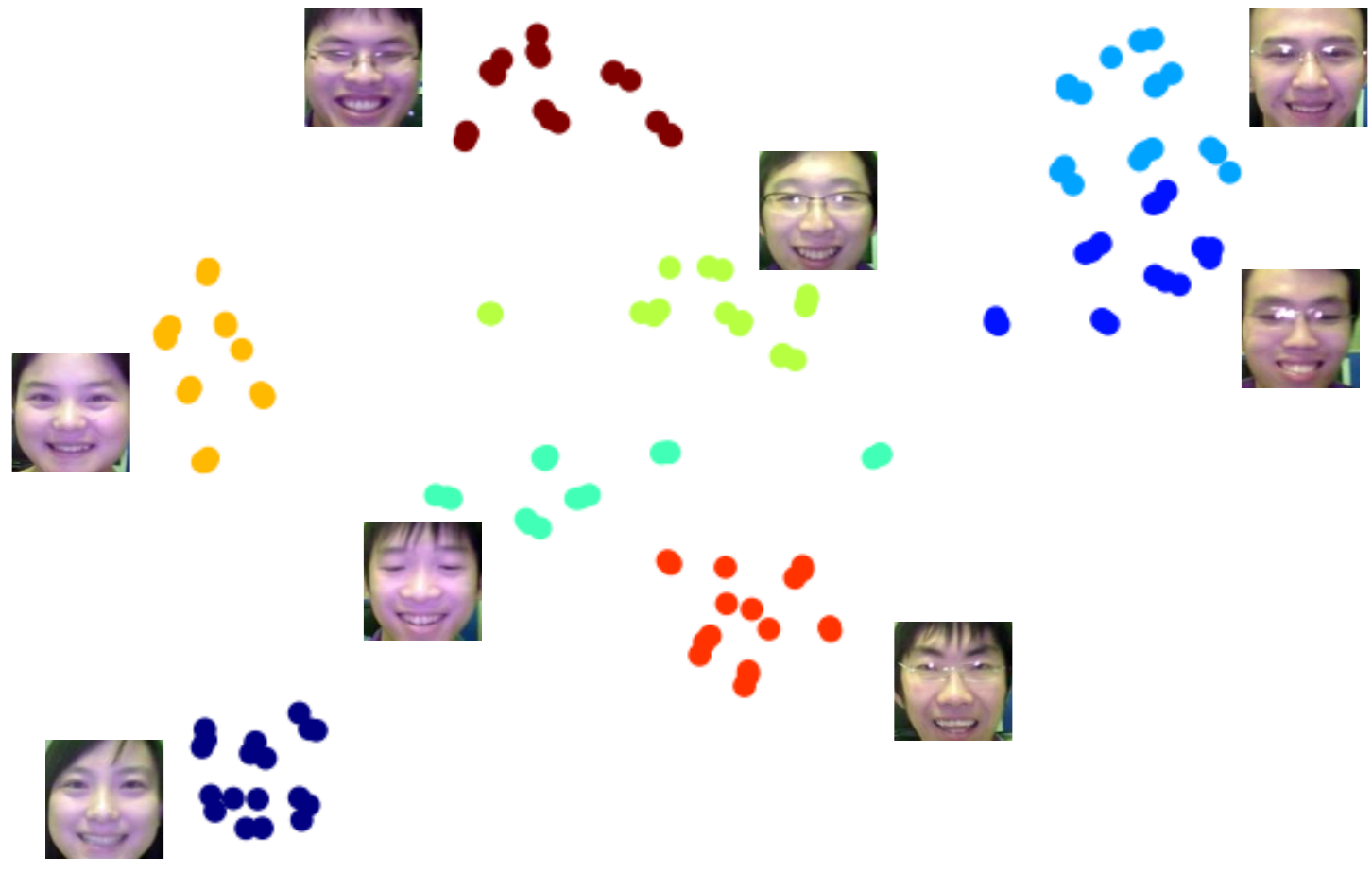}
   \caption{Identity feature space. Each color represents a different identity and the images for one identity are labeled. 
   }
   \label{fig:id_feats}
\end{figure}

In this part, we demonstrate that the identity $g(x)$ and expression $c$ representations learned by our model are disentangled.
To show this, we first use t-SNE~\cite{maaten2008visualizing} to visualize the 50-dim identity feature $g(x)$ on a two dimensional space. The results are shown in Fig.~\ref{fig:id_feats}. 
We can see that most of the subjects are well separated, which confirms the latent identity features $g(x)$ learn to preserve the identity information.

\begin{figure}[!ht]
   \centering
     \includegraphics*[width=0.65\linewidth]{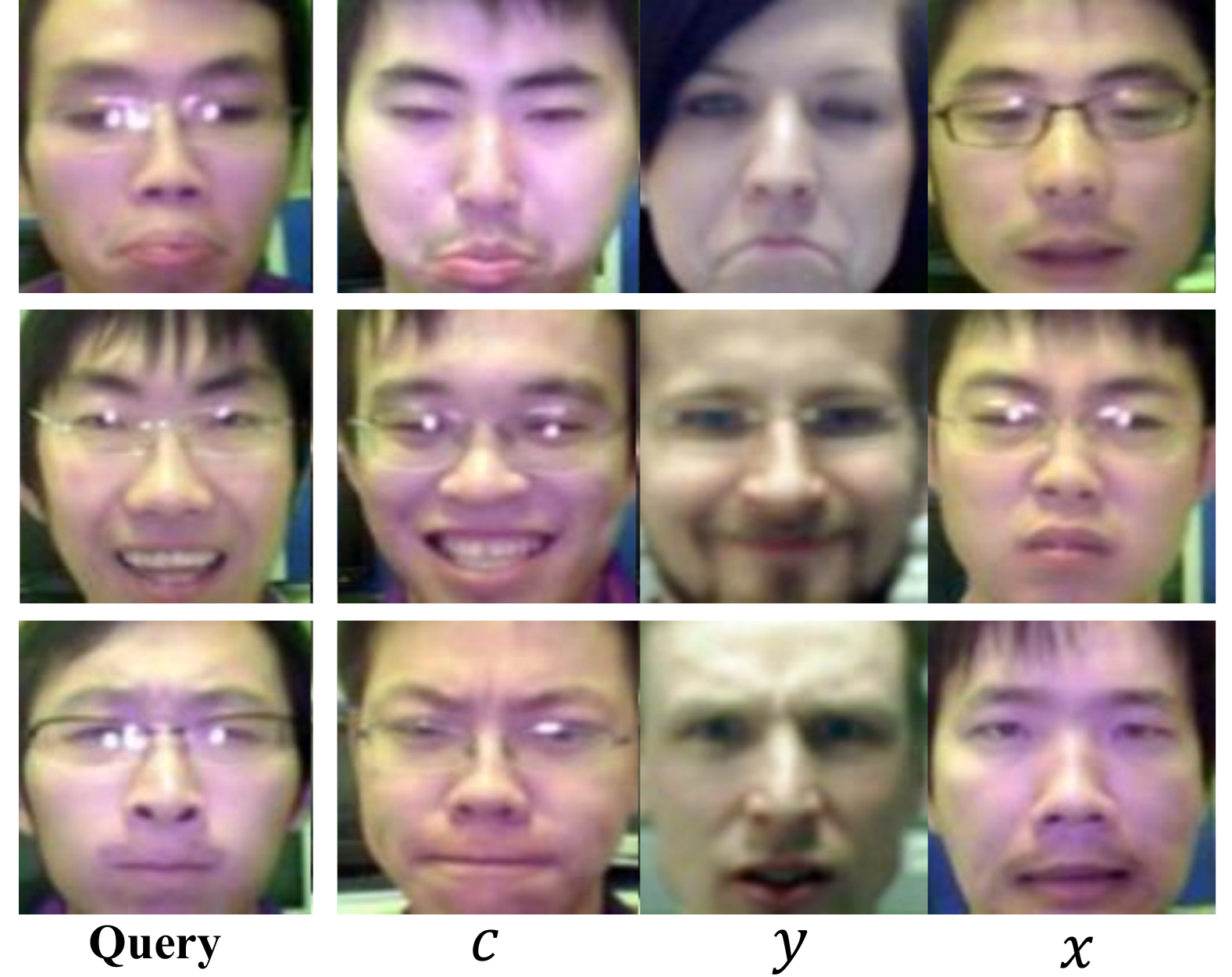}
   \caption{Expression-based image retrieval. First column shows query images. 
   Other columns show top one retrieval based on $c$, $y$ and $x$. 
   }
   \label{fig:expression_feats}
\end{figure}

To demonstrate that the expression code $c$ captures the high-level expression semantics, we perform image retrieval experiment based on $c$ in terms of Euclidean distance. For comparison, the results with expression label $y$ and image pixel space $x$ are also provided in Fig.~\ref{fig:expression_feats}. 
As expected, the pixel space $x$ sometimes fails to retrieve images from the same expression. 
While the images retrieved by $y$ do not always have the same \textit{style} of expressions as the queries. 
For example, the query face in the second row shows a big smile with teeth, but the retrieved image by $y$ only has a mild smile with mouth closed. 
However, with the expression code $c$, 
we observe that face images with similar expressions are always retrieved.
This validates that the expression code learns a rich and diverse feature representation.


\section{Conclusions}
This paper presents ExprGAN for facial expression editing. To the best of our knowledge, it is the first GAN-based model that can transform the face image to a new expression where the expression intensity is allowed to be controlled continuously. 
The proposed model learns the disentangled identity and expression representations explicitly, 
allowing for a wide variety of applications, including expression editing, expression transfer, and data augmentation for training improved face expression recognition models.
We further develop an incremental learning scheme to train the model on small datasets. 
Our future work will explore how to apply ExprGAN to a larger and more unconstrained facial expression dataset.

\bibliographystyle{aaai}
\bibliography{reference}

\end{document}